# A Review of Reinforcement Learning for Natural Language Processing, and Applications in Healthcare

Ying Liu, PhD[1], Haozhu Wang, PhD[2], Huixue Zhou, MS[1], Mingchen Li, MS[1], Yu Hou, PhD[1], Sicheng Zhou, MS[1],

Fang Wang, MS[2], Rama Hoetzlein, PhD[3], Rui Zhang, PhD[1*]

[1] Department of Surgery, University of Minnesota, Minneapolis, Minnesota, USA

[2] Amazon Web Services, Seattle, WA, USA

[3] Quanta Sciences, Ithaca, NY, USA

* Corresponding author: Rui Zhang, PhD

University of Minnesota

zhan1386@umn.edu

## Abstract

Reinforcement learning (RL) has emerged as a powerful approach for tackling complex medical decision-making problems such as treatment planning, personalized medicine, and optimizing the scheduling of surgeries and appointments. It has gained significant attention in the field of Natural Language Processing (NLP) due to its ability to learn optimal strategies for tasks such as dialogue systems, machine translation, and question-answering. This paper presents a review of the RL techniques in NLP, highlighting key advancements, challenges, and applications in healthcare. The review begins by visualizing a roadmap of machine learning and its applications in healthcare. And then it explores the integration of RL with NLP tasks. We examined dialogue systems where RL enables the learning of conversational strategies, RL-based machine translation models, question-answering systems, text summarization, and information extraction. Additionally, ethical considerations and biases in RL-NLP systems are addressed.

## 1 Introduction

Growing healthcare needs driven by population growth, aging, and chronic diseases pose significant challenges. The world population reached 8 billion in 2022 (Zeifman et al., 2022), and healthcare costs are expected to rise substantially over the next 50 years (Tabata, 2005). In this context, innovative solutions are essential to enhance healthcare quality and control costs. ChatGPT (OpenAI, 2022), powered by reinforcement learning (RL) (Sallab et al., 2017), has gained global popularity and holds promise in addressing these challenges. RL, with its capacity to learn from interactions, is well-suited for various applications, including healthcare (C. Yu et al., 2023). Natural Language Processing (NLP) in healthcare has benefited from the widespread adoption of electronic health records (EHR) (Evans, 2016) and the ongoing refinement of the Unified Medical Language System (UMLS) (Lindberg et al., 1993). Building on prior work focusing on RL and general English applications (Uc-Cetina et al., 2023), this paper reviews RL techniques in NLP and emphasizes their applications in biomedical NLP.

To assess the impact of modern technologies on healthcare, we created a Machine Learning (ML), Artificial Intelligence (AI), and Healthcare Development Visualization Map (**Figure 1**) that tracks the evolution of these fields. Starting in the 1980s, the map spans interdisciplinary events and their impacts. We use a logarithmic time scale due to the rapid pace of discovery. We focus on clinical diagnosis, cognitive assistance, genomics, and knowledge work as vertical domains, emphasizing technology's human-based impact on specific patient conditions. The map illustrates connections between medical research and computer science, highlighting the delayed adoption across domains and its subsequent acceleration.

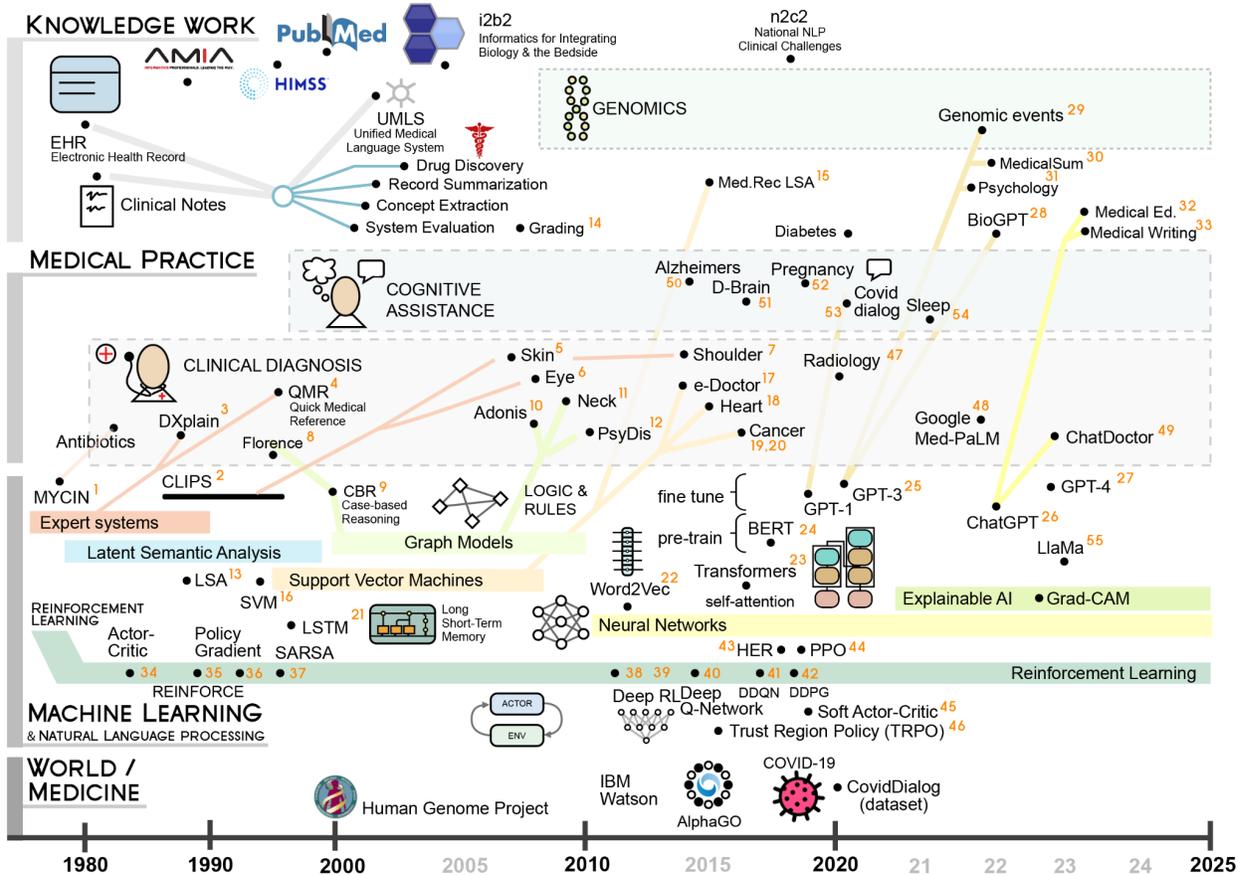

**Figure 1**. A Timeline Map of Machine Learning and NLP Progress in Healthcare. (Citations in the diagram are in Appendix A.)

In clinical diagnosis, ML models can assist in differential diagnosis, where they compare a patient's symptoms and medical data against a vast knowledge base to suggest possible diagnoses (17, Kampouraki et al., 2013; 7, Abu-Naser & Hilles, 2016). In Figure 1, the bottom layer provides the foundation for domain-specific research experiments. For example, in medical imaging analysis, RL algorithms can analyze large amounts of medical images, such as X-rays and CT scans, to assist in detecting abnormalities and generating the reports automatically (47, Nishino et al., 2020). Particularly in NLP, RL algorithms can be applied to a wide range of tasks, such as dialogue systems (49, Yunxiang et al., 2023), machine translation (Lattimore & Szepesvári, 2020), text summarization (Michalopoulos et al., 2022), and question answering (2, Abu-Nasser, 2017). NLP enables the extraction of clinical knowledge from medical literature and guidelines, making it easier for healthcare professionals to diagnose diseases (Cortes & Vapnik, 1995), and suggesting appropriate treatment options (Ling et al., 2017b; Yunxiang et al., 2023). We hope this map might serve as a guide for where to focus future efforts on meaningful work, whether that is advancing new core technology, performing novel medical research, fighting a particular condition, serving a specific group, or directly examining the ways in which technology has impacted medicine.

The rest of the paper was organized in the following way. In section 2, we introduced how we selected the review papers. In section 3, we gave an overview introduction of RL and different applications of RL for NLP. And then, we discussed the ethical concerns in section 4.

## 2 Method

The review adhered to the Preferred Reporting Items for Systematic Reviews and Meta-Analyses (PRISMA) guidelines. A comprehensive search was conducted in multiple databases covering the past 10 years (2014-2023) and limited to English language articles. The databases searched included Ovid MEDLINE(R), PubMed, Scopus, Web of Science, ACM Digital Library, and IEEE Xplore. The final set of keywords used in the search query is presented in **Figure 2.** We selected 89 articles for review (**Figure 3**). The scope of this review spans publications from 11 different countries, with the majority of contributions originating from the United States and China, as illustrated in **Figure 4**.

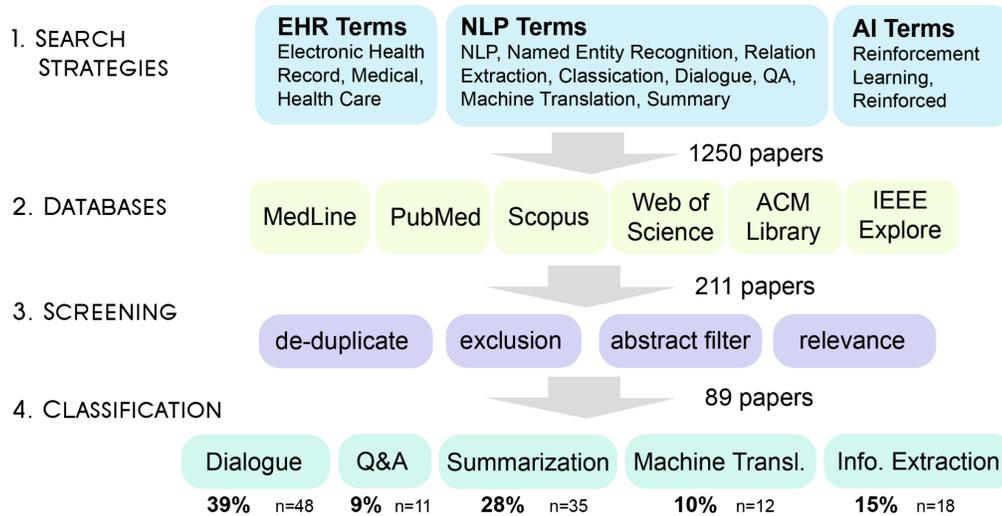

**Figure 2** Searching strategies and Distribution of the reviewed articles.

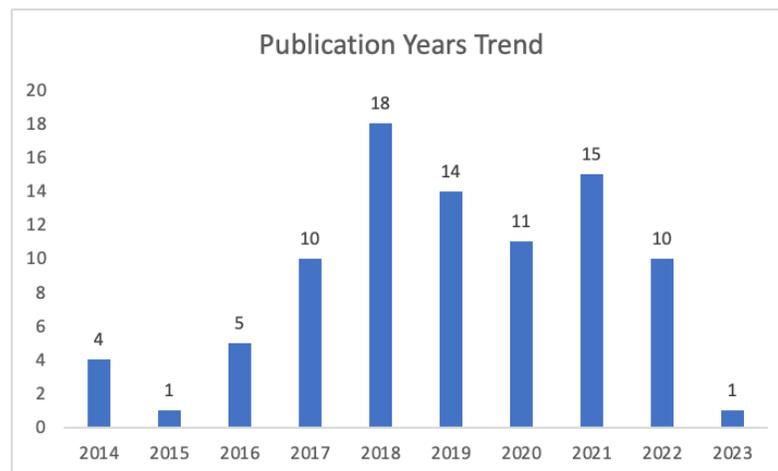

**Figure 3**. Year trend of 89 reviewed studies.

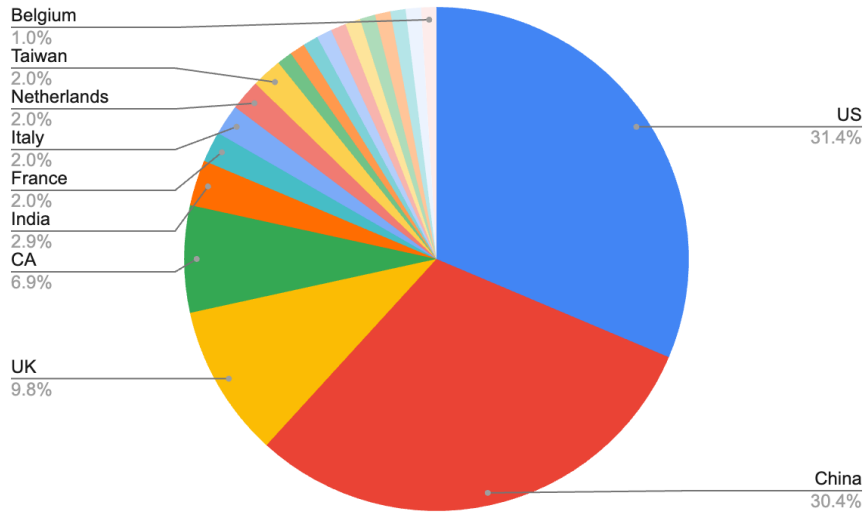

**Figure 4**. Overview of authors' nationalities in the reviewed studies.

## 3 Applied RL and NLP Techniques in Healthcare

### 3.1 Introduction to RL and NLP

Reinforcement Learning (RL) is a machine learning approach where models are trained through a system of rewards and penalties for their actions. RL agents, which are the entities being trained, interact with their environment, learn from their experiences, and aim to achieve specific goals. Unlike supervised learning, RL agents learn through interaction with the environment rather than from labeled data. Key RL concepts include states, actions, and rewards. The agent observes the current state of the environment, takes actions based on that state, and receives rewards in response to its actions. A policy defines how the agent maps states to actions, and the ultimate objective is to find an optimal policy that maximizes cumulative rewards. This learning process is formalized as Markov Decision Processes (MDPs), providing a mathematical framework for modeling sequential decision-making problems (Sutton & Barto, 2018). (**Figure 5**).

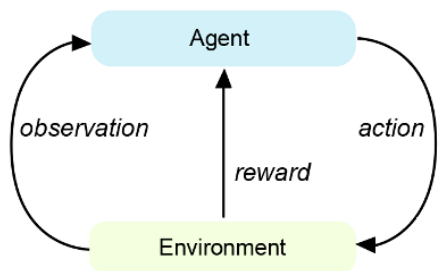

**Figure 5**. MDP process.

According to different environments and applications, RL methods can be classified into on-policy and off-policy models. On-policy methods refer to the process of learning a policy that is used to make decisions during interaction with the environment. For off-policy methods, the agent can learn from experiences generated by a different policy, including random policies. Q-learning is the most commonly used off-policy algorithm that learns the optimal Q-values for each state-action pair by maximizing the expected future reward over all possible actions (Levine et al., 2020). RL algorithms can also be classified into model-based and model-free algorithms. Model-based RL algorithms learn a model of the environment dynamics based on its interactions with the environment. Once the model is learned, the agent can use it to simulate different scenarios and plan its actions accordingly. Model-free RL algorithms, on the other hand, directly learn the optimal policy or action-value function without explicitly building a model of the environment. These algorithms rely on trial-and-error learning, where the agent interacts with the environment, observes the state and reward, and updates its policy or action-value function based on the observed feedback(Sutton & Barto, 2018). **Table 1** listed strength and weakness of these RL methods.

NLP is dedicated to equipping computers with the capability to comprehend, decipher, and even create human language. Some NLP problems can be defined as MDPs Markov Decision Processes (MDPs) and use RL algorithms to solve them in uncertain and sequential environments. NLP tasks, such as question answering, dialogue systems, and machine translation, often involve sequential decision-making processes. In these tasks, the agent needs to make decisions based on the current input or context. The agent takes actions, such as selecting words or phrases, generating responses, or making translations, and receives feedback in the form of rewards or penalties based on the quality of its decisions. In the following sections, we review using the RL algorithm for different applications of NLP.

|  | **Strength** | **Weakness** | **Examples** |
| --- | --- | --- | --- |
| **On-Policy** | 1.Stable<br>2.Built-in exploration | 1. Expensive to collect data.<br>2. Slow convergence and limited policies they can execute.. | Sarsa (State-Action-Reward-State-Action)<br>A2C (Advantage Actor-Critic)<br><br>Robotics<br>Games (AlphaGo, Monte Carlo Tree Search)<br>Auto pilot |
| **Off-Policy** | 1. Flexible because they can learn from any policy.<br>2. Efficient because they can reuse past experiences. | 1.Unstable because they learn from a different policy than the one they are following.<br>2. Bias because they are learning from past experiences. | Q-Learning<br>DDPG (Deep Deterministic Policy Gradient)<br><br>Recommendation systems.<br>Advertising. |
| **Model Based** | 1. Sample Efficient<br>2. Accurate | 1.Computationally expensive<br>2. Model inaccuracy | Dyna-Q<br>Model Predictive Control (MPC)<br><br>Robotics<br>Traffic Control, Auto pilot |
| **Model Free** | 1. Simplicity<br>2. Robustness | 1.Sample inefficient and Exploration-Exploitation | Q-learning<br>Policy Gradient |

| | | trade off because they do not explicitly consider the dynamics of the environment. 2.Lead to suboptimal policy. | Gaming (Q-Learning) Finance (Deep Q-Networks) |
|---|---|---|---|

Table 1, RL methods overview. Weakness and strength of different methods.

For the papers we reviewed, we analyzed various RL techniques. This distribution indicates that policy gradient, deep reinforcement learning, and deep Q networks are the most commonly employed techniques in the NLP applications. GPT-based methods and policy networks also hold significant prominence. Additionally, actor-critic, hierarchical RL, and several other techniques contribute to the research efforts, albeit to a slightly lesser extent. These findings highlight the diverse array of RL algorithms being investigated to address NLP challenges (**Figure 6**).

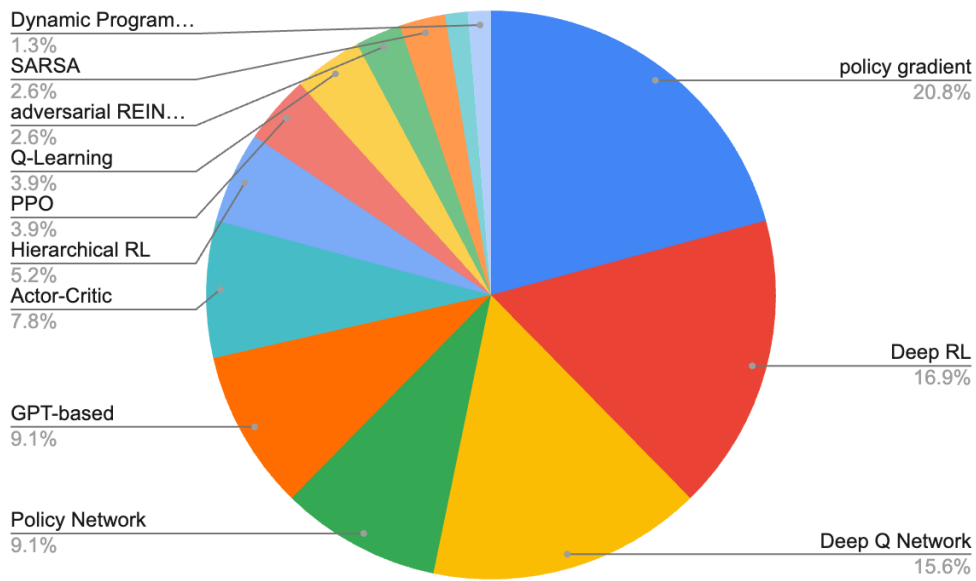

Figure 6. RL algorithms distribution used in the review papers.

## 3.2 Dialogue

There are two primary classifications of dialogue systems: open domain (J. Li et al., 2017; Z. Yu et al., 2016; Zhou et al., 2019) and task-oriented systems (Rajendran et al., 2018; T. Zhao & Eskenazi, 2016). Task-oriented chatbots can be designed to assist users in customer service (Eshghi et al., 2017; Peng et al., 2017; Rieser et al., 2014; Shi et al., 2019; Wei et al., 2018), personal assistants (Korpusik & Glass, 2019; M.-H. Su et al., 2016), and e-commerce (Sun & Zhang, 2018). The key feature of task-oriented dialogue systems is their ability to understand the user's intent and take actions accordingly. This requires the system to be trained on a specific domain (Budzianowski et al., 2017; Takanobu, Zhu, et al., 2019; Z. Wang et al., 2014) to have a knowledge base and actions. Building an effective task-oriented dialogue system requires large amounts of training data to respond to a wide range of user inputs to solve the large state-action space challenge (Kwan et al., 2023). However, the collection of vast amounts of data may be challenged in various cases. Low-data training (Siddique et al., 2022; P.-H. Su et al., 2017) techniques, such as transfer learning (Ilievski et al., 2018) and data augmentation, are used to leverage existing data and create more diverse training data. Transfer learning involves training the system on a large dataset from a related domain, and then fine-tuning it on the target domain with the limited available data. Data augmentation involves generating new training data from the existing data by applying various transformations, such as changing word order or synonym replacement. User simulators (Kreyssig et al., 2018; Shi et al., 2019; Tseng et al., 2021) can provide a way to generate realistic user inputs and evaluate the performance of the system in a controlled environment.They are also often used for the development and testing of dialogue systems.

Modern chatbots use RL techniques with large amounts of training data to improve their performance. RL's reward mechanism can help the dialogue system to improve quality (Kandasamy et al., 2017), decision making (Kreyssig et al., 2018; Shi et al., 2022), and gives diverse answers (Chou & Hsueh, 2019). ChatGPT has become popular all over the world due to its advanced capabilities to engage in conversation with humans, multiple languages capability, and continuously adapt to changing user needs and preferences (Ray, 2023). ChatGPT is a large language model trained on a diverse range of textual data using deep learning techniques. Reinforcement learning with human feedback (RLHF) has played an essential role in helping ChatGPT to become a more effective and efficient language model (Ouyang et al., 2022). Proximal Policy Optimization (PPO) is regarded as one of the leading approaches in RL (Y. Wang et al., 2019). This algorithm was introduced by OpenAI in 2017 and is considered to be a good compromise between performance and comprehension. PPO is particularly well-suited for dialogue systems as it addresses some of the challenges, such as dealing with sparse and delayed rewards, handling large action spaces, and managing the exploration-exploitation trade-off in a dialogue setting.

Medical dialogue systems hold great potential to aid telemedicine by enhancing the accessibility of healthcare services, elevating the standard of patient care, and decreasing medical expenses. (Shim et al., 2021) investigated three aspects of task-oriented dialogue systems. Firstly, data privacy is a critical issue in any system that deals with personal information. Secondly, medical knowledge is the fundamental requirement for a healthcare dialogue system.Thirdly, it is crucial for the dialogue system to undergo a human-centered evaluation. (Wei et al., 2018) created a dataset using real conversations between doctors and patients and demonstrated a dialogue system that can learn information about new symptoms from patient conversations, leading to improved accuracy in automated diagnoses. (Chinaei & Chaib-draa, 2014) used unannotated data collected at SmartWheeler (Pineau et al., 2010) to learn dialogue strategy by RL. By offering guidance and support to expecting mothers throughout their pregnancy, an intelligent bot system acting as a healthcare expert has the potential to bolster efforts aimed at reducing maternal mortality rates (Mugoye et al., 2019). During the pandemic period of Covid-19, in order to solve the problem of insufficient medical staff and too many patients needing consultation issues, (G. Zeng et al., n.d.) and (Tripathy et al., 2023) developed the Covid consulting dialogue consulting systems. In recent years, with the development of deep learning and large language models, (G. Zeng et al., 2020) created the MedDialog database and included transformer (BertGPT) and GPT-2 based models. ChatDoctor (Yunxiang et al., 2023) is a medical dialogue system based on fine-turn large language models (LLMs) to tailor to the medical domain.

### 3.3 Question and Answering

A question answering (QA) system is an AI system that responds to human queries by providing relevant answers. QA systems can be designed for customer service, technical support, or educational purposes. They can also be used to assist with research and analysis, by quickly retrieving relevant information from large volumes of text. IBM Watson was a leading QA platform, and it demonstrated its ability to successfully answer a wide variety of questions in real-time and won the Jeopardy Champion in 2011 (High, 2012). (Li, M., & Ji, S. 2022) developed a Semantic Structure-based Query Graph Prediction system for Question Answering over Knowledge Graphs. This system has been evaluated and proven capable of identifying reasonable answers while providing interpretable query paths. Now, the most influential dialogue and QA platform is ChatGPT. To answer a question, ChatGPT first processes the input question through its neural network and generates a contextualized representation of the question, and then it searches through its knowledge base. Once relevant information has been identified, it uses language generation capabilities to produce a response to the question (OpenAI, 2022).

A QA system in the healthcare domain is designed to provide accurate and reliable answers to questions related to medical information, diagnoses, or treatments. Key Components of a QA System in the medical domain include data collection, information retrieval, question understanding, answer generation, confidence estimation, and continuous learning and updates (Budler et al., 2023). QA systems require a comprehensive collection of medical literature, research papers, clinical guidelines, and electronic health records. QA systems involve indexing and ranking of documents based on relevance to the query. IBM Watson for Oncology is a renowned QA system that uses cognitive computing capabilities to assist oncologists in making treatment decisions for cancer patients. It analyzes patient data, medical literature, and clinical guidelines to provide personalized treatment options and recommendations (Strickland, 2019). Buoy Health is an AI-driven digital health platform that offers a symptom checker and triage service. Users can describe their symptoms, and the system uses NLP techniques to understand the user's intent, provide potential diagnoses, and offer appropriate healthcare recommendations, such as self-care advice or the need for medical consultation (Roy & Baksi, 2022). RL helped to ask the right medical questions by adaptive feature selection and applied it to a national survey data to predict 4-year mortality of patients (Shaham et al., 2020). The intelligent QA system for Tibetan medicine, which is based on a knowledge graph (KG), utilizes RL for optimizing feedback (Dong et al., 2021). Another medical consultant system (Huang et al., 2021) used RL to help patients answer healthcare questions. ChatDoctor (Yunxiang et al., 2023) is a new QA system based on large language models. Another example of a LLM is Google's Med-PaLM (Medical Patient Language Model), which is designed to assist clinicians in their daily practice (Singhal et al., 2022). PubMedQA (Q. Jin et al., 2019) is a benchmark dataset to test the QA system in the medical domain. Healthcare professionals can potentially improve efficiency, accuracy, and overall patient care. However, it's important to note that LLMs are meant to be used as tools to support clinical decision-making, and the final responsibility and decision-making authority still rest with the healthcare providers.

Accuracy is one of the most important factors for a successful QA system with ground truth answers. RL can help learn to ask the right questions and clarify any misunderstandings (Buck et al., 2018). This helps clarify the user's intent and correct errors. For example, on retrieval-based QA systems, (Buck et al., 2018) generates the complete question reformulations by the traditional seq2seq model, and then the token-level F1 score is computed as the reward for each reformulation question. (B. Liu et al., 2019) combines three rewards to reorganize the ill-formed question and improve the quality of generated answers. RL can also help ranking the answers. (S. Wang et al., 2018) proposed a pipeline which ranked retrieved information by using RL and likelihood. Another key factor to consider is the stability of a QA system. (M. Zhang et al., 2019) set up a home service-oriented QA system by using RL to design the rewards to provide accurate and stable responses to user questions. The system also demonstrates high stability by maintaining consistent performance across multiple test runs. Other methods include used multi-document summarization techniques to perform complex question answering and formulate it as a RL problem (Chali et al., 2015), using QA to interactively learn language (Yuan et al., 2019)

Many QA systems were built on knowledge graphs (KG). RL can help to construct the knowledge base (Dong et al., 2021; Huang et al., 2021; Q. Zhang et al., 2022). For complex questions which need more than one step to get the answers, (Qiu et al., 2020) proposes a hierarchical reinforcement learning method to reward each element in the query graph and train the model by weak supervision. This model solved the multi-relation question issue as a sequential decision process by using a stepwise reasoning network. To tackle the potential distribution bias in questions, (Hua et al., 2020) employs a retriever to find similar questions in the training data, and then a good programming policy which utilizes the trial trajectories and their rewards for similar questions in the support set.

## 3.4 Machine Translation

Machine Translation (MT) employs algorithms to convert text or speech between languages. MT has evolved from rule-based and statistical techniques to advanced methods like neural networks. Ensuring translation accuracy and fluency is a challenge. Feedback from humans, often through bandit feedback, improves translation quality. Bandit feedback can be seen as a single-state Markov Decision Process (MDP) with multiple actions, aiding fast convergence (Lattimore & Szepesvári, 2020; K. Nguyen et al., 2017).

Approximately 25.6 million limited English proficient (LEP) individuals in the USA are affected by language barriers (Khoong & Rodriguez, 2022). Machine translation can help LEP patients by providing them with communication with doctors. However, use of MT in the healthcare domain still has risks because one error translation could cause harmful consequences. An assessing study about Google translation, 2% of Spanish and 8% of Chinese sentence translations had potential for significant harm (Khoong et al., 2019). Reliability and safety are crucial for MT in the healthcare domain (Mehandru et al., 2022). Because of the cost, many hospitals do not offer professional medical interpreters for verbal communication, and it is unlikely to offer translation of text instructions. MT aids efficient patient-provider communication but carries risks. It may not grasp complex medical terms or concepts, leading to mistranslation or information loss. Cultural and linguistic differences can also impede effective communication in medical settings.

Re-phasing can be considered as a MT problem. To facilitate empathetic conversations in online mental health support systems, Sharma et al (A. Sharma et al., 2021) developed a RL system to rewrite sentences from a lower empathetic tone to a higher empathetic tone which helped improve engagement with patients. Wu et al (L. Wu et al., 2018) demonstrates how to train Neural machine translation (NMT) with RL effectively. They found multinomial sampling is better than beam search and that using MLE and RL together helped to improve the quality of translation. (Geng et al., 2018) utilized RL to construct a multi-pass decoder that manages the decoding depth to improve translation quality. To solve the delayed translation problem, RL can be used for word prediction. Word prediction helps with simultaneous translation and helps improve translation quality (Alinejad et al., 2018; Grissom Ii et al., 2014) . Tebbifakhr et al. (Kumari et al., 2021; Tebbifakhr et al., 2019, 2020) used machine oriented quality criteria , selecting the best translation by suiting a downstream NLP task. NLP components are fed with translations produced by a single NMT system, which is adapted to generate output that is "easy to process" by the downstream processing tools.

## 3.5 Text Summarization and Simplification

Automatic text summarization (ATS) is the process of reducing a larger text document to a shorter version while retaining the most important information. There are two main approaches: extractive and abstractive. Extractive summarization involves selecting important sentences or phrases from the original text and presenting them as a summary (Narayan et al., 2018; Y. Wu & Hu, 2018). Abstractive summarization involves generating new sentences that capture the meaning of the original text (Alomari et al., 2022; Paulus et al., 2017). Q. Wang et al., 2019) created a Bert based hybrid model by bridging extractive-abstractive approaches via a RL method. This method is more challenging, as it requires a deep understanding of the text and the ability to generate new sentences that are both grammatically correct and semantically meaningful.

RL-based models represent the state of the art algorithms. They work by maximizing a reward signal, which can be defined as a function of the quality of the generated summary, such as the ROUGE score (C.-Y. Lin, 2004). ROUGE looks for exact matches between references and generations, but it fails to capture the semantic relation between similar words. The alternative ways are directly compared with human feedback (Scialom et al., 2019; Stiennon et al., 2020), or other metrics that measure the similarity between the generated summary and a reference summary (Böhm et al., 2019; Gigioli et al., 2018; Jang & Kim, 2021; S. Li et al., 2019; Rioux et al., 2014).

Summarization can be used for news summarization (Grenander et al., 2019), headline generation (Erraki et al., 2020; Gigioli et al., 2018), and legal documents (D.-H. Nguyen et al., 2021). Not limited to text, summarization can also be used for data (Trummer, 2022) and source code (Wan et al., 2018). (Z. Li et al., 2020) proposed an encoder-decoder model based on a double attention pointer network (DAPT). In DAPT, the self-attention mechanism collects key information from the encoder, the soft attention and the pointer network generate more coherent core content, and the fusion of both generates accurate and coherent summaries. (Gao et al., 2018) used active preference learning and RL to summarize. (E. Sharma et al., 2019; Tian et al., 2019) 2-step system by first retrieving the entities or word type, and then generating the final word distribution based on the predicted type. (Gigioli et al., 2018) proposed an abstractive summarization model capable of reading biomedical publication abstracts and producing one-sentence titles. They introduced deep RL reward metrics based on biomedical expert tools such as the UMLS Metathesaurus and MeSH and demonstrated that it can produce domain-aware summaries.

Text simplification involves converting intricate texts into simpler and more easily understandable language. This process plays a crucial role in assisting individuals with learning disabilities, non-native speakers, and anyone who may struggle with comprehending complex language. The reward signal is typically based on a combination of factors such as grammaticality, fluency, and simplicity (X. Zhang & Lapata, 2017). Text simplification is evaluated using metrics such as Flesch-Kincaid Grade Level (Kincaid et al., 1975) or Simple Measure of Gobbledygook (SMOG) (Grabeel et al., 2018). Text simplification for medical information involves techniques such as replacing complex medical terms with simpler alternatives, breaking down complex sentences into simpler ones, using plain language, and adding visual aids such as diagrams and illustrations. (Phatak et al., 2022) simplified medical text at paragraph level, with the goal to enhance the accessibility of biomedical research to a wider audience.

## 3.6 Information Extraction

Information Extraction (IE) autonomously organizes insights from unstructured text data, identifying entities like names, locations, and connections. SemRep is a rule-based biomedical relation extraction (RE) tool for PubMed abstracts (Kilicoglu et al., 2020) While rule-based methods are precise and interpretable, creating rules can be labor-intensive. Most recently, (Li M, Ye Y, Yeung J, et al., 2023) introduced Weighted Prototypical Contrastive Learning as a solution to address class collision issues in few-shot medical Named Entity Recognition (NER). Additionally, (Li, M., & Zhang, R. 2023) proposed a novel approach that leverages in-context learning and a chain-of-thought technique to enhance the performance of medical NER.

RL-based NER and RE approaches offer advantages: learning from unlabeled text, adaptability to new domains, and potential discovery of novel relationships. However, they demand significant computational resources and may lack interpretability. (B. Liu et al., 2019) addressed these challenges using DNN-based RE models, incorporating human knowledge as soft rules and relation evidence.

Traditional NLP tasks separate RE and NER, but researchers are increasingly using RL methods to tackle them together (Takanobu, Zhang, et al., 2019; W. Zhao et al., 2021; Zhu & Zhu, 2021;Li, M., & Huang, L.2023). These problems can be addressed at the sentence or document level. At the sentence level, RL, bidirectional LSTM, and attention mechanisms solve relation extraction (Z. Liu et al., 2021). Cleaning noisy labels is vital for accuracy (Feng et al., 2018; Z. Liu et al., 2021), particularly in clinical notes(L. Xu et al., 2020). Combining RL with Deep Q Network (DQN) (T. Chen et al., 2020) or neural networks (Qin et al., 2018) addresses noise. Document-level relation extraction is more complex as it spans multiple sentences or documents. RL can use graph structures to infer relationships between entities mentioned in different sections (T. Xu et al., 2022). Word vectors with co-occurrence statistics or semantic similarities assist in such inferences (X. Zeng et al., 2018).

In the biomedical domain, sentences contain complex structures and a wide range of domain-specific terminology, including acronyms and technical jargon, which make it difficult to identify the correct entities and relations. Zhao et al (W. Zhao et al., 2021) used RL to extract multiple events and relations by trigger identification and argument detection as main and subsidiary tasks respectively. Unbalanced data sets are common for biomedical text. Combining RL and data augmentation, such as synonyms/class and word position replacement, can further enhance this ability to obtain data features (A. Wang et al., 2022). (Camara et al., 2015) developed a multi-agent system for distributed text mining of biomedical information. (Ling et al., 2017a) used RL to find the most probable diagnoses by optimizing clinical concept extraction.

## 4 Discussion

Traditional supervised NLP models are effective for tasks like NER and RE, but they struggle with complex linguistic structures and lack adaptability. In contrast, neural network-based models like RNNs, CNNs, and transformers excel at capturing complex semantics, achieving state-of-the-art performance. RL-based algorithms, such as ChatGPT, are ideal for sequential decision-making and interaction in NLP, accommodating user preferences and evolving language generation. They explore diverse strategies and adapt to changing environments. However, RL faces challenges, including substantial data requirements, sample inefficiency, exploration dilemmas, and handling sparse or delayed rewards. Adapting to non-stationary language environments while ensuring safe and responsible behavior remains a complex challenge in applying RL to NLP tasks.

While RL in NLP for healthcare is in its early stages, the potential for transformative solutions is vast. Medical dialogue systems are a promising application, enhancing telemedicine accessibility and reducing costs. These systems aid expecting mothers, support military personnel and families, and address healthcare staff shortages, but ensuring patient safety is paramount. QA systems in healthcare, like IBM Watson for Oncology and Buoy Health, provide evidence-based answers, but human oversight remains crucial. Machine translation aids communication in healthcare but faces challenges with medical terminology and data privacy. Text summarization and IE benefit from RL but require large, specialized medical text datasets. RL-NLP in healthcare has potential but demands attention to privacy, quality, interpretability, and ethics to ensure responsible and effective use.

**Figure 7** illustrates the advantages and weaknesses we need to consider for RL-NLP in healthcare applications. In the reviewed five major domains, RL in NLP for healthcare applications has the potential to enhance healthcare accessibility, support clinical decision-making, and improve patient outcomes. However, it also faces challenges related to data privacy, quality assurance, interpretability, and ethical considerations etc. Addressing these issues is crucial to ensure the responsible and effective use of RL in healthcare NLP.

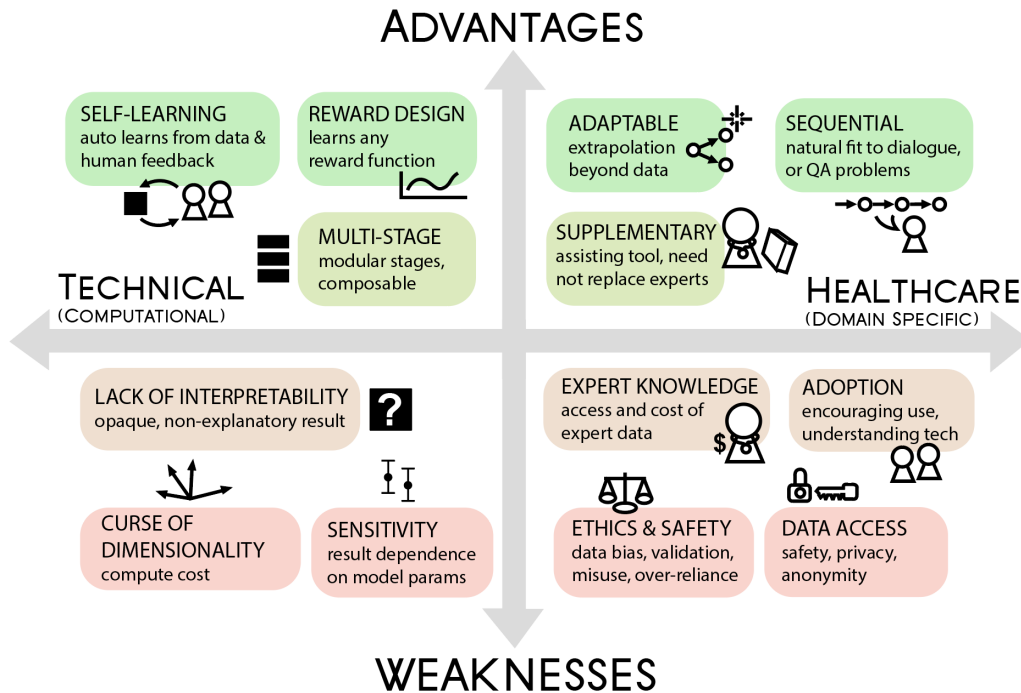

**Figure 7.** The advantages and weaknesses of RL to consider for healthcare applications.

Reinforcement learning in model training raises ethical concerns. It relies on rewards and penalties, potentially leading to biased behavior or sensationalist content generation. RL's black-box nature makes decision-making and behavior factors unclear, raising accountability and harming potential worries. If not properly trained or in poorly designed environments, RL can result in unintended consequences, such as clickbait or spam. The substantial data and computational power needed for training LLMs also raise privacy and environmental concerns.

# 5 Author Contributions

YL wrote the paper. HW and FW contributed to the organization of the paper. HZ reviewed the Dialogue section. ML reviewed the Machine Translation section. YH reviewed the Information Extracted section. Sicheng reviewed the Summarization section. RH did the diagrams. RZ is the corresponding author and was responsible for the overall progress.

# 6 Acknowledgement

This work was supported by the National Institutes of Health under grant number 2R01AT009457, 1R01AG078154, 1R01CA287413, and 1R21MD019134. The content is solely the responsibility of the authors and does not represent the official views of the National Institutes of Health.

## Appendix A:

References as indicated on the healthcare map (Figure 1)

1. (Shortliffe & Buchanan, 1975) 2. (Abu-Nasser, 2017) 3. (Barnett et al., 1987) 4. (Rassinoux et al., 1996) 5. (Naser & Akkila, 2008) 6. (Samy & Zaiter, 2008) 7. (Abu-Naser & Hilles, 2016) 8. (Bradburn & Zeleznikow, 1994) 9. (Schmidt et al., 2001) 10. (Rodriguez et al., 2009) 11. (Abu-Naser & Almurshidi, 2016) 12. (Casado-Lumbreras et al., 2012) 13. (Deerwester et al., 1988) 14. (Kintsch, 2002) 15. (X. Jin et al., 2015) 16. (Ahsan et al., 2022; Cortes & Vapnik, 1995) 17. (Kampouraki et al., 2013) 18. (Singh et al., 2019) 19. (Abdel-Nasser et al., 2015) 20 (*Predicting Breast Cancer via Supervised Machine Learning Methods on Class Imbalanced Data - ProQuest*, n.d.) 21. (Hochreiter & Schmidhuber, 1997) 22. (Mikolov et al., 2013) 23. (Vaswani et al., 2017) 24. (Devlin et al., 2019) 25. (Brown et al., 2020) 26. (OpenAI, 2022) 27. (OpenAI, 2023) 28. (Luo et al., 2022) 29. (W. Zhao et al., 2021) 30. (Michalopoulos et al., 2022) 31. (Binz & Schulz, 2023) 32. (Kung et al., 2023) 33. (Biswas, 2023) 34. (Barto et al., 1983) 35. (Williams, 1992) 36. (Sutton et al., 1999) 37. (Rummery & Niranjan, 1994) 38. (Silver et al., 2016) 39. (Sutton & Barto, 2018) 40. (Mnih et al., 2015) 41. (Z. Wang et al., 2016) 42. (Lillicrap et al., 2015) 43. (Andrychowicz et al., 2018) 44. (Schulman, Wolski, et al., 2017) 45. (Haarnoja et al., 2018) 46. (Schulman, Levine, et al., 2017) 47. (Nishino et al., 2020) 48. (Singhal et al., 2023) 49. (Yunxiang et al., 2023) 50. (Chinaei & Chaib-draa, 2014) 51. (Ling et al., 2017b) 52. (Mugoye et al., 2019) 53. (Yang et al., 2020) 54. (Shim et al., 2021) 55.(Touvron et al).